\documentclass[conference]{IEEEtran}
\usepackage{times}
\usepackage{graphics} 
\usepackage{graphicx}
\usepackage{amsmath}
\usepackage[numbers]{natbib}
\usepackage{multicol}
\usepackage{amssymb}
\usepackage{xcolor}
\usepackage[export]{adjustbox}
\usepackage[bookmarks=true]{hyperref}

\begin{document}

\title{Interactive Plan Explicability in\\ Human-Robot Teaming}

\author{Mehrdad Zakershahrak and Yu Zhang\\
Computer Science and Engineering Department\\
Arizona State University\\
Tempe, Arizona \\
mzakersh, yzhan442@asu.edu}

\maketitle

\begin{abstract}
Human-robot teaming is one of the most important applications of artificial intelligence in the fast-growing field of robotics. For effective teaming, a robot must not only maintain a behavioral model of its human teammates to project the team status, but also be aware that its human teammates' expectation of itself. Being aware of the human teammates' expectation leads to robot behaviors that better align with human expectation, thus facilitating more efficient and potentially safer teams. 
Our work addresses the problem of human-robot cooperation with the consideration of such teammate models in sequential domains by leveraging the concept of plan explicability. 
In plan explicability, however, the human is considered solely as an observer.
In this paper, we extend plan explicability to consider interactive settings where human and robot behaviors can influence each other. 
We term this new measure as 
Interactive Plan Explicability. 
We compare the joint plan generated with the consideration of this measure using the fast forward planner (FF) with the plan created by FF without such consideration, as well as the plan created with actual human subjects. 
Results indicate that the explicability score of plans generated by our algorithm is comparable to the human plan, and better than the plan created by FF without considering the measure,
implying that the plans created by our algorithms align better with
expected joint plans of the human during execution.
This can lead to more efficient collaboration in practice.
\end{abstract}

\IEEEpeerreviewmaketitle

\section{Introduction}
The notion of a robotic teammate,
or that using robots to complement humans in various tasks, 
has attracted a lot of research interest. 
At the same time, the realization of this notion is challenging due to the human-aware aspect \cite{chakraborti2017ai},
or that the robot must consider the human in the loop, in terms of both
physical and mental models
while achieving the team goal. 
In such cases, it is no longer sufficient to model humans passively as parts of the environment \cite{chakraborti2015planning}.
Instead, human-robot teaming applications require the robot to be proactive in assisting humans \cite{fern2007decision}.

There are different aspects to be considered for human-robot teaming. First, the robot must take the human's intent into account. 
Various plan recognition algorithms \cite{kautz1986generalized,ramirez2010probabilistic} can be applied to perform plan recognition based on a given set of observations. The challenge is how the robot can utilize this information to synthesize a plan while avoiding conflicts or providing proactive assistance \cite{chakraborti2016planning, cirillo2009human}. There are different approaches to planning with such consideration \cite{chakraborti2015planning, chakraborti2017plan}. Another the key consideration is to be socially acceptable \cite{dragan2013generating,zhang2016plan}, where the robot must be aware of expectation of the human teammates and acts accordingly. The challenge here is to model the human's expectation of the robot. 

The ability to model the human's expectations enables
the robot to assist humans in an expected and understandable fashion that is consistent with the teaming context \cite{knepper2017implicit}. This type of coordination results in effective teaming \cite{cooke2015team}. One of the key challenges for such effective teaming is for the robot to learn the human's preconceptions about its own model, as illustrated in Figure \ref{fig1}. To learn about this model, similar to \cite{zhang2017plan}, we assume that humans understand other agents' behavior by associating abstract tasks with agent's actions. Alternatively, when the robot's behavior does not match that of the human's expectation, the human would not be able to associate some of its actions with task labels. The labeling process can be learned using conditional random fields (CRFs). Then, the learned model can be used to label a new robot plan to compute its explicability score. The explicability measure in \citet{zhang2017plan} is defined as follows:

\textit{Plan Explicability}: After a plan is labeled, its explicability score is computed based on its action labels. The explicability score is calculated as follows:
\begin{equation}
F_\theta(L_\pi) = \dfrac{\sum_{i \in [1,N]}1_{L(a_i) \neq \emptyset}}{N} \label{score}
\end{equation}
where $F_\theta(L_\pi)$ : $L_\pi \rightarrow [0,1]$ (with 1 being the most explicable), $\pi$ is the robot plan, 1 is an indicator function, $N$ is the total number of actions in the plan, and $L_\pi$ denotes the sequence of action labels for plan $\pi$, and $F_{\theta}$ is a domain independent function that converts plan labels to the final score. When the labeling process can't assign a label to an action $a_i$, its label $L(a_i)$ will be empty.

In this work, we extend the notion of plan explicability to an interactive setting where the human is cooperating with robot. In such a case, a plan is comprised of both human and robot actions, and the influence of the agent's behavior on each other must be explicitly considered. Another contribution is the implementation and evaluation of our approach in a first response task domain in simulation.



\section{Interactive Plan Explicability}
The explicability of a plan \cite{zhang2016plan} is correlated with a mapping of high-level tasks (as interpreted by humans) to the actions performed by the robotic agent. The demand for generating explicable plans is due to the inconsistencies between the robot's model and the human's interpretation of the robot model \cite{zakershahrak2018interactive}. In our work, the robot creates composite plans for both the human and robot using an estimated human model and the robot's model, which can be considered as its prediction of the joint plan that the team is going to perform. At the same time, however, the human would also anticipate such a plan to achieve the same task, except with an estimated robot model and the human's own model.

\begin{figure}[t]
\centering
\includegraphics[scale=.35]{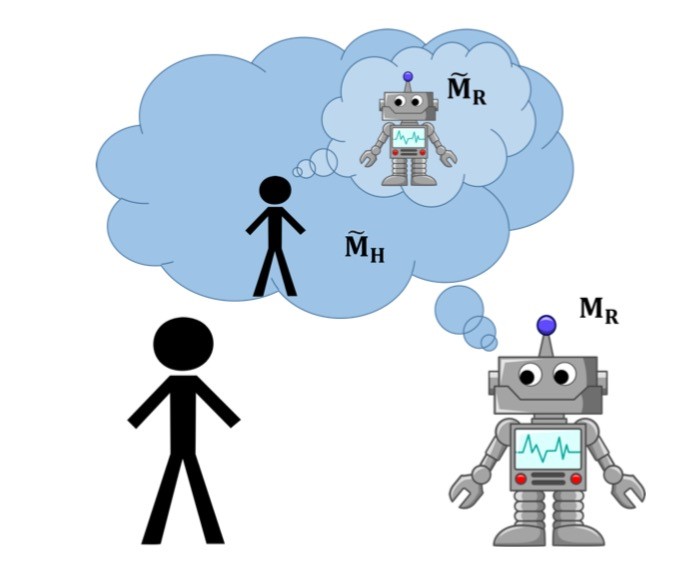}
\caption{The robot's planning process is informed by an approximate human planning model as well as the robot's planning model.}
\label{fig1}
\end{figure}

Each problem in this domain can be expressed as a tuple $P_T = \langle I, G, M_R, \widetilde{M_H}, \Pi_C \rangle$. In this tuple, $I$ denotes the initial state of the planning problem, while $G$ represents the shared goal of the team. $M_R$ represents the actual robot model and $\widetilde{M_H}$ denotes the approximate human planning model provided to the robot. The actual human planning model $M_H$ (that the human uses to create his own prediction of the joint plan) could be quite different from the model $\widetilde{M_H}$ provided to the robot. Similarly, the human will be using $\widetilde{M_R}$ that may be different from the actual robot model $M_R$. Finally $\Pi_C$ represents a set of annotated plans that are provided as the training set for the CRF model.

To generate an explicable plan,
the robot needs to synthesize a composite plan that is as close as possible to the plan that the human expects. This is an especially daunting challenge, given that we have multiple points of domain uncertainty (e.g. from $\widetilde{M_H}$ and $\widetilde{M_R}$). As shown in Figure \ref{fig1}, the robot only has access to $\widetilde{M_H}$ and $M_R$. Thus, the problem of generating explicable pan can be formulated as the following optimization problem:
\begin{multline}
argmin_{\pi_C^{M_R, \widetilde{M_H}}} cost (\pi_C^{M_R, \widetilde{M_H}}) \\ 
+ \alpha \cdot dist(\pi_C^{M_R, \widetilde{M_H}}, \pi_C^{\widetilde{M_R}, M_H})
\end{multline}
where $\pi_C^{M_R, \widetilde{M_H}}$ is the composite plan created by the robot using $M_R$ and $\widetilde{M_H}$, while $\pi_C^{\widetilde{M_R}, M_H}$ is the composite plan that is assumed to be created by the human (the plan that the human expects). 
Similar to \cite{zhang2016plan}, we assume that the distance function $dist(\pi_C^{M_R, \widetilde{M_H}}, \pi_C^{\widetilde{M_R}, M_H})$ can be calculated as a function of labels of actions in $\pi_C^{M_R, \widetilde{M_H}}$.
\begin{multline}
argmin_{\pi_C^{M_R, \widetilde{M_H}}} cost (\pi_C^{M_R, \widetilde{M_H}}) \\ 
+ \alpha \cdot F \circ L_{CRF}(\pi_C^{M_R, \widetilde{M_H}} \mid \{S_i \mid S_i = L^{\star} (\pi_C)\}) \label{ab}
\end{multline}
 
As shown in (\ref{ab}), the label for each action is produced by a CRF model $L_{CRF}$ trained on a set of labeled team execution traces ($\pi_C)$. Since we do not have access to the human model or the human's expectation of the robot model so that mispredictions are expected, we will rely on replanning when either the human deviates from the predicted plan of the robot.

To search for an explicable plan, we use a heuristic search method, $f = g+h$, where $g$ is the cost of the plan prefix and $h$ is calculated as shown in the following:
\begin{equation} \label{heu}
    h = (1.0-F_{\theta }(L(state.path \# rp))) * |state.path \# rp| * |rp| + |rp|
\end{equation}
where $\#$ means concatenation above and $rp = relaxedPlan(state, Goal)$.


\section{Evaluation}
To evaluate our system, we tested it on a simulated first response domain, where a human-robot team is assigned to a first-response task after a disaster occurred. In this scenario, the human's task is to team up with a remote robot that is working on the disaster scene. The team goal is to search all the marked locations as fast as possible and the human's role is to help the robot by providing high-level guidance as to which marked location to visit next. The human peer has access to the floor plan of the scene before the disaster. However, some paths may be blocked due to the disaster that the human may not know about; the robot, however, can use its sensors to detect these changes. Due to these changes in the environment, the robot might not take the expected paths of the human.

For data collection, we implemented the discussed scenario by developing an interactive web application using MEAN (Mongo-Express-Angular-Node) stack. 

In our setting,
the robot would always follow the human's command (i.e., which room to visit next). 
The human can, of course, change the next room to be visited by the robot anytime during the task if necessary, simply by clicking on any of the marked locations. The robot uses BFS search to plan to visit the next room. After a room is visited, the human cannot click on the room anymore. Also, the robot always waits 1 second before performing the next action. For simplicity, the costs of all human and robot actions are the same.

 \subsection{Experimental Setup} For training,
 after each robot action, the system asks the human whether the robot's action makes sense or not. If the human answers positively, that action is considered to be explicable. Otherwise, the action is considered to be inexplicable. This is used later as the labels for learning the model of interactive plan explicability. 
All scenarios were limited to four marked locations to be visited, with a random number ($2-5$) of visible obstacles and manually inserted hidden obstacles (invisible to the human) in the map. 
We have generated a set of 16 problems for training and 4 problems for testing. 


We collected in total 34 plan traces for training, which were used to train our CRF model. All training data was collected with human trials, with random initial robot initial and goal locations.  To remove the influence of symbol permutation, we performed the following processing on the training set: For each problem, we created an additional $1000$ traces that are the same problem only with different permutations of symbols. 

\begin{figure} [t]
\centering
\includegraphics[scale =0.4]{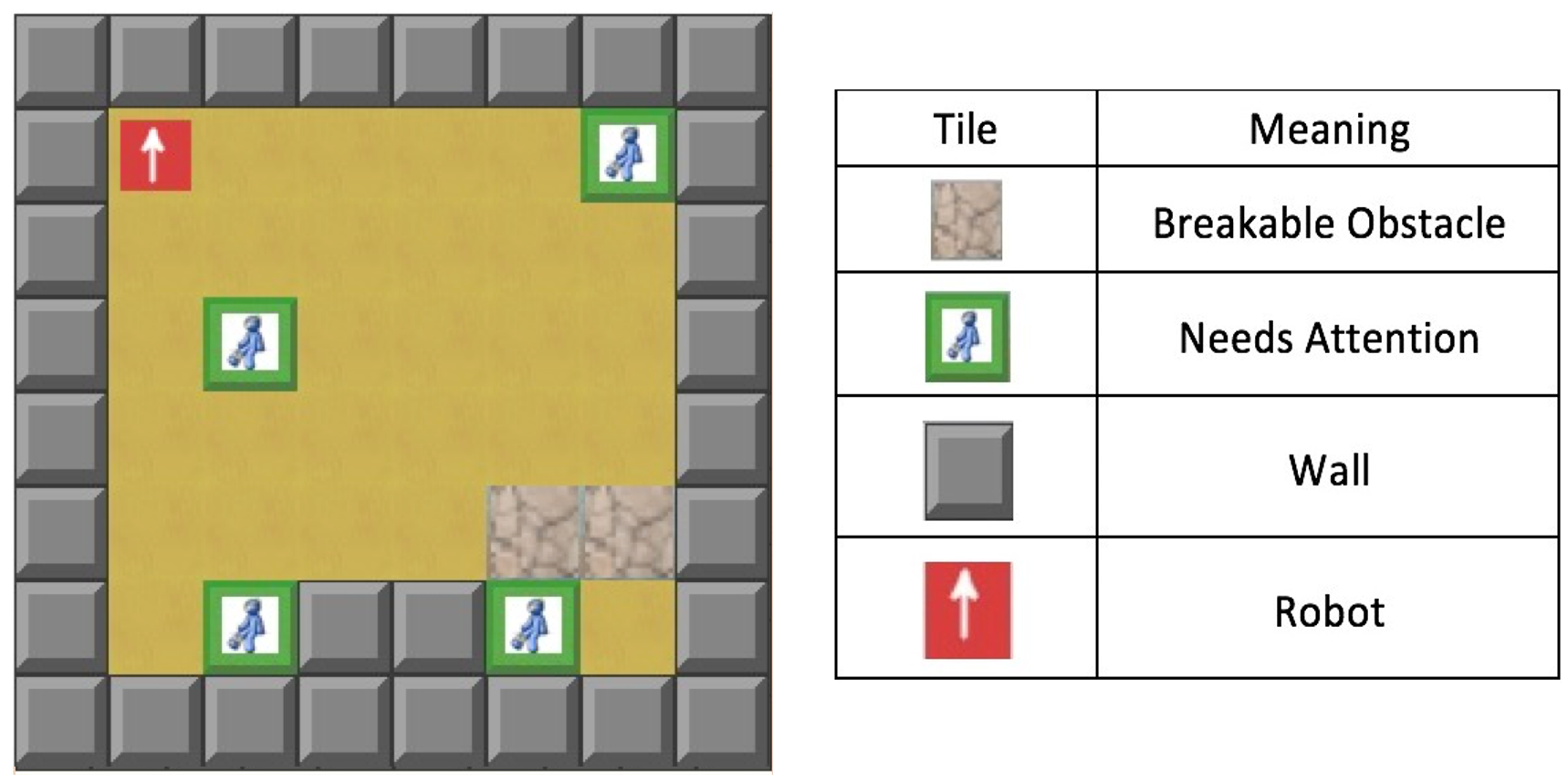}
\caption{ A sample map that the human subjects see with a description of the object types.}
\label{tile}
\end{figure}

\begin{figure} [t]
\centering
\includegraphics[scale =0.36]{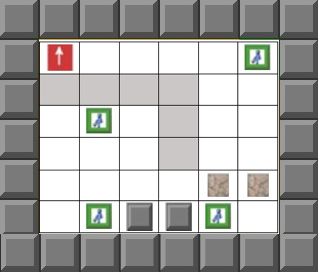}
\caption{A sample map corresponding to the map in Figure \ref{tile} that the robot sees; the gray cells are hidden obstacles.}
\label{robotv}
\end{figure}

A sample map of the actual environment is shown in Figure \ref{tile}. Figure \ref{robotv} shows the same map that the robot sees with hidden obstacles drawn on the map.

\subsection{RESULTS}
Table \ref{result} shows the ratios (refer to as the {\it explicability ratio}) between the number of explicable actions and the number of actions over all plans, created for the testing problems using our approach, FF planner, and human plan, respectively. 
 The interactive explicable plan (our approach) is created using the heuristic search method mentioned in Equation (\ref{heu}).
Note that all the human actions will be considered explicable in our plans (although one can argue that is not the case).

\begin{figure*} [ht]
\centering
\includegraphics[scale =0.22]{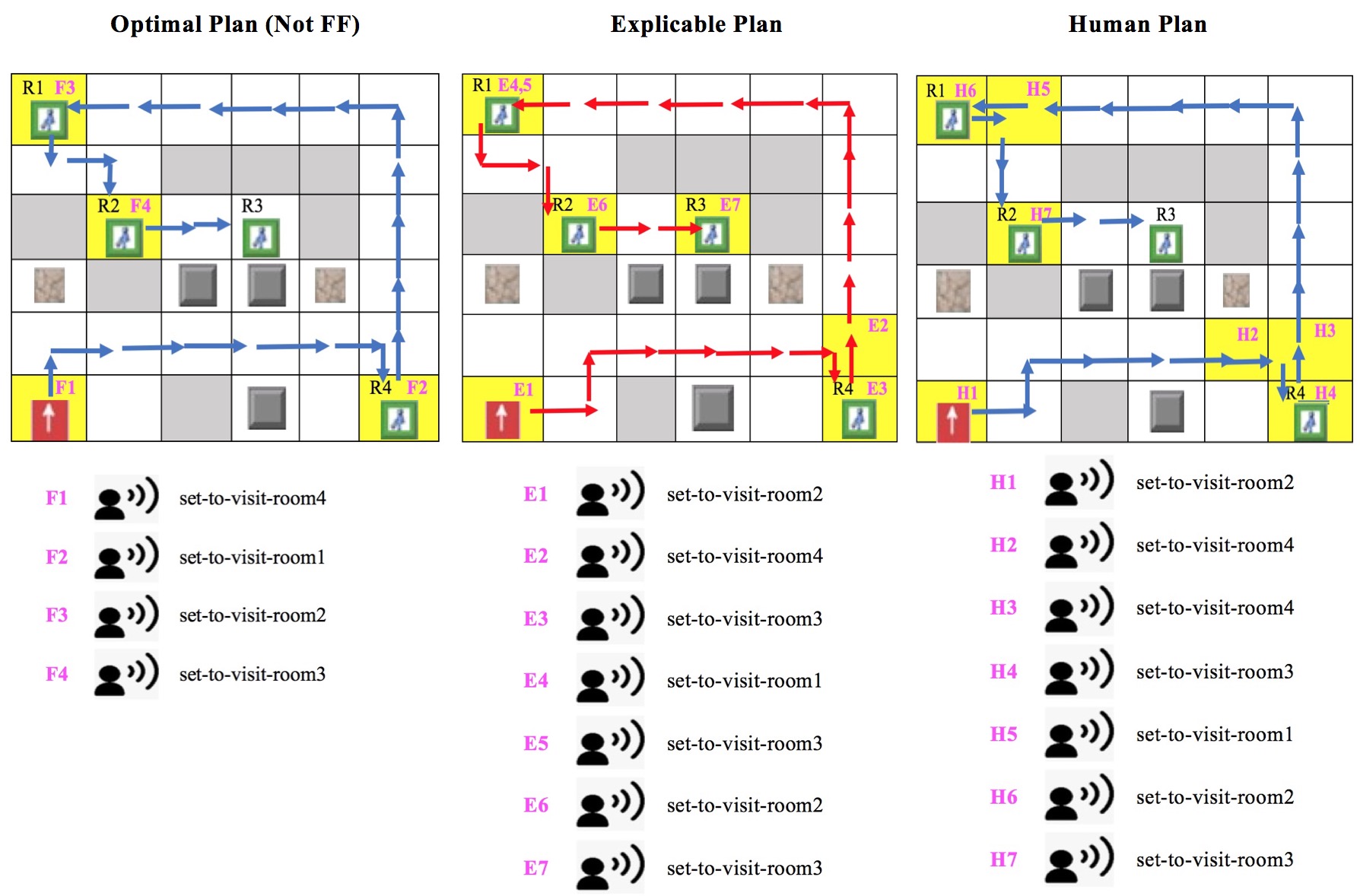}
\caption{Comparison of plans for a specific problem. (Left) The optimal plan; (Middle) The explicable Plan; (Right) The human plan. The initial location of the robot is indicated with a white arrow inside a red box. Yellow cells refers to where the human commands are received.
}
\label{fig3}
\end{figure*}

 As we can see in Figure \ref{fig3}, the explicability ratio for our approach is similar (0.1\% difference) to the human plan while being quite different from the FF plan (13.9\% difference).
 This is also intuitively explained in Fig. \ref{fig3}, where
 We can clearly see that the explicable plan is similar to the human plan, in the sense the human tends to change commands in this task domain due to unknown situation. 

The above results show that the plans created by our algorithm are closer to what the human expects, and thus enabling the robot to better predict the team behavior and potentially lead to more efficient collaboration in practice.
The explicability scores for the four testing problems are shown in Table \ref{actions}.
The reason for the low explicability score of FF plan is that FF tends to create plans that are less costly while ignoring the fact that the human and robot may view the environment {\textbf\textit{and each other}} differently, and thus less costly plans in one view are also more likely to be misaligned with less costly plans in the other. 
Note, however, that whether the explicable plan would
lead to better teaming performance (e.g., less replanning efforts for the robot and less cognitive load for the human) requires further investigation and evaluation with actual human subjects.
This will be explored in future work.
\begin{table}[ht]
\caption{Comparison of Explicability Ratio For Testing Scenarios}
\label{result}
\begin{center}
\begin{tabular}{|l|c|}
    \hline
    Plan Type            & Interactive Explicability Score \\ \hline
    Interactive Explicable Plan          & 0.820                 \\ \hline
    FF Planner & 0.672                \\ \hline
    Human Plan           & 0.811                \\ \hline
    \end{tabular}
\end{center}
\end{table}
\begin{table}[h]
\caption{Elaborated explicability Score for Test Scenarios}
\label{actions}
\begin{center}
\begin{tabular}{|c|c|c|}
    \hline
    Scenario \# & FF Plan & Interactive Explicable Plan \\ \hline
    1           & 1.0   & 1.0              \\ \hline
    2           & 0.56   & 0.714              \\ \hline
    3           & 0.629   & 0.757              \\ \hline
    4           & 0.8    & 0.8             \\ \hline
    \end{tabular}
\end{center}
\end{table}

\section{CONCLUSIONS AND FUTURE WORK}

We created a general way of generating explicable plans for human-robot teams, where the human is an active player.
This differs from prior work in the sense that we do not assume that the human and robot have the same knowledge about the environment and each other; or in other words, there exists information asymmetry, which is often true in realistic task domains.
To generate an explicable plan for a human-robot team, we need not only consider the plan cost, but also the preconceptions that the human may have about the robot.
Although we have mainly focused on two member teams, we believe that these ideas can be easily extended to larger team sizes with a few changes to the current formulation. It should also be straightforward to extend the current formulation to support simultaneous action executions by considering joint actions at any time step. Another way we may be able to achieve this would be by using temporal planners \cite{do2003sapa} instead of relying on sequential ones. Also, the current system assumes the provision of an approximate human planning model and relies on replanning to correct its plans whenever the human deviates from the predicted explicable plan. We could possibly explore the idea of incorporating models like capability model \cite{zhang2015capability} to learn such human models.



\bibliographystyle{plainnat}
\bibliography{references}

\end{document}